\documentclass[reprint,amsmath,amssymb,aps]{revtex4-2}

\usepackage{graphicx}
\usepackage{dcolumn}
\usepackage{bm}
\usepackage[table,x11names]{xcolor}
\usepackage{physics}
\usepackage{amsmath}
\usepackage[colorlinks=true, allcolors=blue]{hyperref}
\usepackage{amssymb}
\usepackage{mathtools}

\newcommand{\be}{\begin{equation}}
\newcommand{\bea}{\begin{eqnarray}}
\newcommand{\ee}{\end{equation}}
\newcommand{\eea}{\end{eqnarray}}

\newcommand{\bes}{\begin{equation*}}
\newcommand{\beas}{\begin{eqnarray*}}
\newcommand{\ees}{\end{equation*}}
\newcommand{\eeas}{\end{eqnarray*}}

\def\r{\mbox{{\bf r}}}

\def\v{\mbox{$\bf{v}$}}

\def\u{\mbox{$\bf{u}$}}

\def\F{\mbox{$\bf{F}$}}

\begin{document}

\title{\Large Forecasting Using Reservoir Computing:\\The Role of Generalized Synchronization}

\author{Jason A. Platt}
\email{jplatt@ucsd.edu}
\author{Adrian Wong}
\email{asw012@ucsd.edu}
\author{Randall Clark}
\email{r2clark@ucsd.edu}
\affiliation{Department of Physics,\\
University of California San Diego\\
9500 Gilman Drive\\
La Jolla, CA 92093\\}

\author{Stephen G. Penny}
\email{steve.penny@noaa.gov}
\affiliation{
Cooperative Institute for Research in Environmental Sciences\\
at the University of Colorado Boulder, and\\
NOAA Physical Sciences Laboratory\\
Boulder, CO, 80305-3328, USA \\}

\author{Henry D. I. Abarbanel}
\email{habarbanel@gmail.com}
\affiliation{
Department of Physics,\\
and\\
Marine Physical Laboratory,\\
Scripps Institution of Oceanography,\\
University of California San Diego\\
9500 Gilman Drive\\
La Jolla, CA 92093}

\date{\today}

\begin{abstract}
Reservoir computers (RC) are a form of recurrent neural network (RNN) used for forecasting time series data.  As with all RNNs, selecting the hyperparameters presents a challenge when training on new inputs. We present a method based on generalized synchronization (GS) that gives direction in designing and evaluating the architecture and hyperparameters of a RC. The `auxiliary method' for detecting GS provides a pre-training test that guides hyperparameter selection.  Furthermore, we provide a metric for a ``well trained'' RC using the reproduction of the input system's {\it Lyapunov exponents}.
\end{abstract}
\maketitle
\section{Introduction}
Machine learning (ML) is a computing paradigm for data-driven prediction in which an ML ``device'' accepts input data in a training phase, which is then used in a predict/forecast phase that is used to extrapolate to new data.  When the data is in the form of a time series, such a ``device'' is denoted a ``recurrent neural network'' (RNN)~\cite{goodfellow16}. 
This is in contrast to other ML network architectures, such as feed forward neural networks, that assume the statistical independence of inputs~\cite{Lipton15,ty19}.

RNNs have feedback in the connection topology of the network, enabling self excitation as a dynamical system and distinguishing them from feed forward networks that only represent functions ~\cite{Luko09}.  This feature identifies RNNs as an attractive choice for data driven forecasting~\cite{Vlachas20}.

A kind of RNN architecture with demonstrated capability for dynamical systems forecasting is reservoir computing (RC)~\cite{Luko09,Verstraeten09,Luko12,
Jaeger01, Jaeger12, Jaeger04, Schrauwen07, Maass02, Wojcik04,pathak18}, where a large random network is constructed and only the final layer is trained. This method is much simpler to train due to the fixed weights in the reservoir layer.  The $D-$dimensional training input signal to the network, denoted $\u(t) \in \mathbb{R}^D$, may be generated from a known dynamical system ~\cite{Hunt19,ottdresden19}, or from observations where the underlying dynamical rules are undetermined. RC's simplicity and ease of training give them a clear advantage over other RNNs.

The ability to develop a data-driven model using a method such as RC is attractive for a number of practical reasons. RC allows us to construct predictive models of unknown or poorly understood dynamics. Should the input signal $\u(t)$ arise from measurements of high dimensional geophysical or laboratory flows~\cite{Sharan19,Matheou18}, the speedup in computing with a reservoir network realized in hardware~\cite{Tanaka19,canaday18} may permit the exploration of detailed statistical questions about the observations that might be difficult or impossible otherwise. RC has the potential to provide significant computational cost savings in prediction applications, since the RC dynamics typically comprises a network with computationally simple active dynamics at its nodes. 

\begin{figure}[!htpb]
    \centering
    \includegraphics[width = 0.49\textwidth]{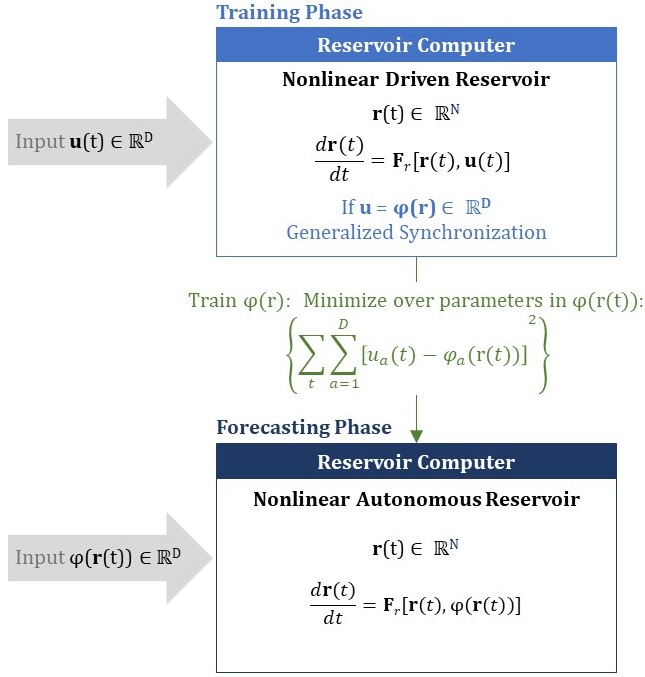}
    \caption{Flow of operations for utilizing a Reservoir Computation (RC) to perform forecasting/prediction of a D-dimensional input $\u(t)$ presented to an RC with N-dimensional dynamical degrees-of-freedom $\r(t)$. When the input and the reservoir exhibit {\it generalized synchronization}, $u_a = \varphi_a(\r); a = 1,2,...D$, training consists of estimating any parameters in a representation of $\varphi(\r)$. }
    \label{fig: res_pic}
\end{figure}

The success of RNNs and their increased adoption in research applications has rapidly outpaced the understanding of these data driven processes. It is not known how best to design a network for a particular problem, nor how much or what kind of data is most useful for training. General guidelines are well established~\cite{Luko12}, but tend to be justified with empirical rather than theoretical considerations. In probing this question, the idea arose ~\cite{Lu18,Hunt19,Lymburn19} that the explanation might be a form of synchronization known as `generalized synchronization' (GS) ~\cite{Abarbanel95,Abarbanel96,Kocarev96}.

We use this insight to move from an {\it ad hoc} training approach to a systematic strategy where we ensure that the input $\u(t)$ and the reservoir degrees-of freedom $\r(t)$ show {\bf generalized synchronization} $u_a(t) = \varphi_a(\r(t));\;a=1,2,...,D$, and point out that it is parameters in the function $\varphi(\r)$ that we need to estimate. We show a computationally efficient way to choose a region of RC hyperparameters---including the spectral radius (SR) of the adjacency matrix $A_{\alpha,\beta}$ and the probability of non-zero connections among the N active units (pnz)---where generalized synchronization occurs and skillful forecasting of the training input data $\u(t)$ is to be expected, given an accurate enough approximation to $\varphi(\r)$. We argue that the vague references in the literature to the ``edge of chaos'' are not particularly informative.

A related issue is that the traditional method of evaluating the effectiveness of an RNN, with training and testing data sets, is not sufficient when performing dynamical systems forecasting, as the former method gives no indication of the stability of the forecast.  A simple approach to evaluation, showing a prediction of a single time series, also gives no indication of the stability of the predictions over the entire range of inputs.  In this paper we attempt to rectify these deficiencies by using dynamical properties of the reservoir to design and evaluate a trained network.

The goals of this paper are as follows:
\begin{itemize}
\item Introduce a numerical test, based on GS and using the `auxiliary method', that can guide hyperparameter selection in RNNs.
\item Provide a metric for a ``well trained'' network using the reproduction of the input system's Lyapunov exponent spectrum.
\end{itemize}

\section{Reservoir Computing}

RCs are applied to forecasting problems where the task is to predict a $D$-dimensional input sequence $\vb u(t)$ generated from a dynamical system 
\bes
    \frac{d\u(t)}{dt} = \F_u(\u(t)), 
    \label{eq: input}
\ees
$\F_u(\u(t))$ is the vector field of the $\u$ dynamics. 

An RC consists of three layers: an input layer, the reservoir itself, and an output layer.  The reservoir, described as a dynamical system $\F_r$, is composed of $N$ nodes at which we locate nonlinear models, in ML called `activation functions',~\cite{Wojcik04, Brunner16, Chrisantha03,Haynes14,Larger17,Paquot11}.  The nodes in the network are connected through an $N \times N$ adjacency matrix $A_{\alpha\,\beta}$, chosen randomly to have a connection density $pnz$ and non-zero elements uniformly chosen from a uniform distribution in $[-1, 1]$. This is then normalized by the largest eigenvalue of $A_{\alpha\,\beta}$: the spectral radius (SR).

\begin{figure}[!htpb]
\centering
\includegraphics[width=0.5\textwidth]{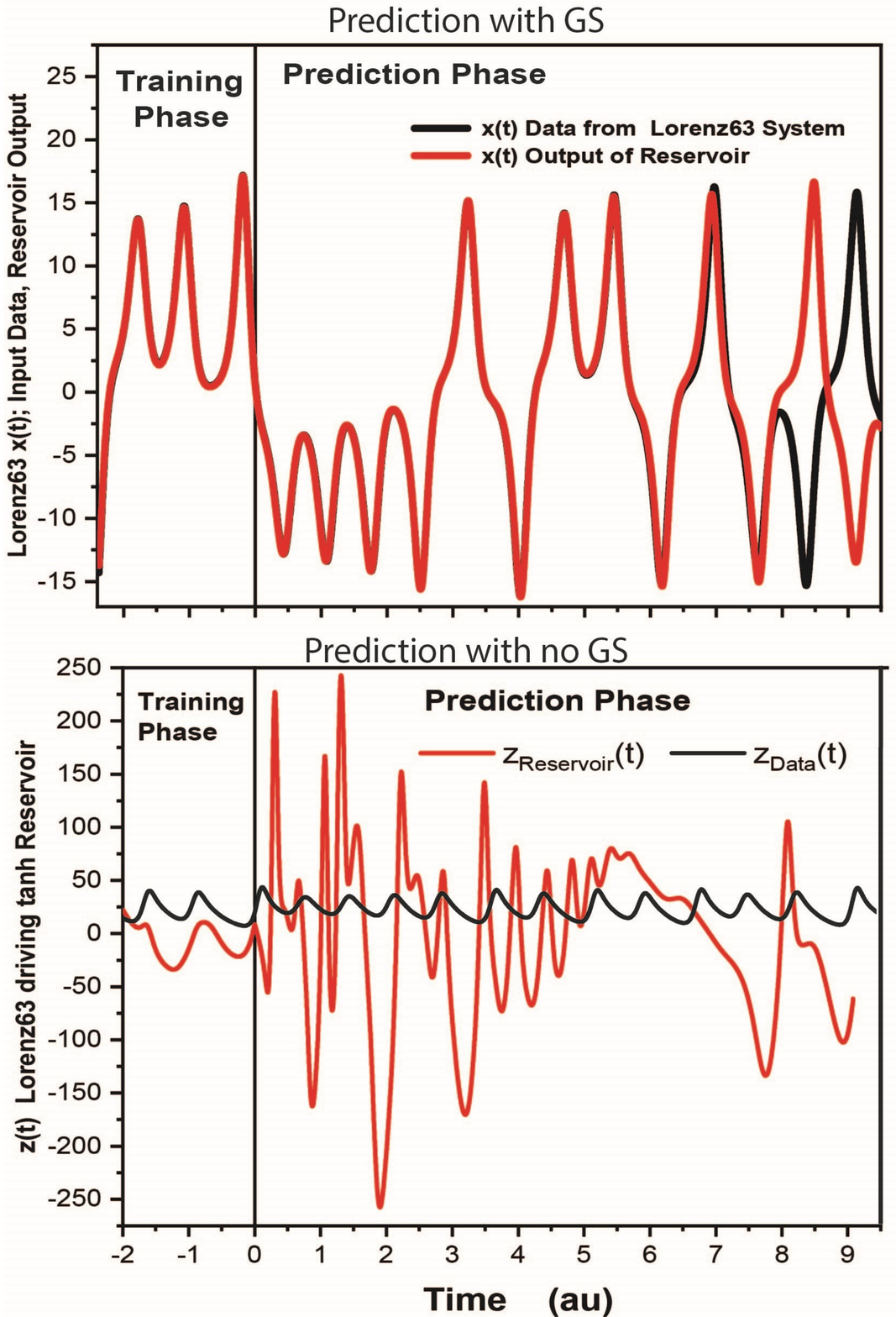}
    \caption{\\{\bf Top}
    Synchronization and prediction between an $N = 2000$ $\tanh$ reservoir output (red) and the Lorenz63 input (black) \cite{Lorenz63, Platt_PRE}. In $A_{\alpha\,\beta}:$ SR = 0.9 and pnz = 0.02.  
    The black vertical line at t = 0 is the end of the ``training period.''\\ {\bf Bottom} When one selects the hyperparameters {\bf outside} the region of GS, for example using $N = 2000$, $SR = 1.6$ and $pnz = 0.02$ for the $\tanh$ reservoir, the function $\varphi(\r)$ does not exist. We may expect the reservoir to operate poorly in producing a replica of the input $\u(t)$.}
    \label{xlor63tores}
\end{figure}

The input layer maps the input signal $\u(t)$ from $D$ dimensions into the $N$ dimensional reservoir space. The output layer $\varphi(\r(t))$ is a function such that $\varphi_a(\r)= u_a(t),$ chosen during the {\bf training phase} during which we estimate any parameters in $\varphi(\r)$. This is the only part of the reservoir computer that is trained.  It is common practice to choose $\varphi(\r)$ as a linear function of $\r$, but this is by no means the only choice of output function \cite{Platt_PRE}.

The structure of a RC is shown in Fig.(\ref{fig: res_pic}).  $\r(t)$ can be viewed as representing the information in the input time series  $\{\u(0), \u(1),\ldots,\u(t_{final})\}$ in $N > D$ dimensional space consistent with Takens' embedding theorem~\cite{Takens81}.

The reservoir dynamics act at the nodes of the network $\r(t)$.  In the `training phase' and `prediction/forecast phase' the equations governing the dynamics of the reservoir are these:
\be
    \underbrace{\frac{ d\r(t)}{dt} = \F_r[\r(t), \u(t)]}_{\rm{training}} \overbrace{\Rightarrow}^{\vb u(t) = \varphi(\vb r(t))} \underbrace{\frac{d\r(t)}{dt} = \F_r[\r(t),\varphi(\r(t))]}_{\rm{forecast}}.
    \label{eq: synchronization_pred}
\ee
The forecast phase in Eq.\eqref{eq: synchronization_pred} is an autonomous dynamical system, enabling prediction.

\subsection{The Hyperbolic Tangent Model}

We use linear operator $\bf{A_{\alpha\,\beta}}$ and nonlinear activation function $tanh()$ \cite{Lu18}. We also use a scaling constant $\gamma$ to adjust the timescale of the reservoir dynamics, and a scaling constant $\sigma$ to weight the influence of the input signal $\u(t)$
\bes
\frac{d\r_\alpha (t)}{dt} = \gamma \{-r_{\alpha}(t) +\tanh(A_{\alpha \beta}r_{\beta}(t) + \sigma W_{\alpha a}u_a(t))\}.
\ees
Repeated indices are summed over. This is not the only formulation that is possible for $\F_r$, see \cite{Platt_PRE} for other kinds of reservoirs including those based on nonlinear neuron models.

\subsection{Synchronization and Training}
GS refers to the synchronization of two {\bf nonidentical} dynamical systems.  These cannot exhibit identical oscillations~\cite{Abarbanel95,Kocarev96,Pyragas98}. For an input time series $\u(t)$ and response system $\r(t)$ GS means there is a function $\varphi(\r)$ connecting them so we have $u_a(t) = \varphi_a(\r(t))$.  We do not have an explicit form for $\varphi$, but if there is GS then we infer it exists~\cite{Abarbanel96,hunt97}.  

\begin{figure}
    \centering
    \includegraphics[width=0.5\textwidth]{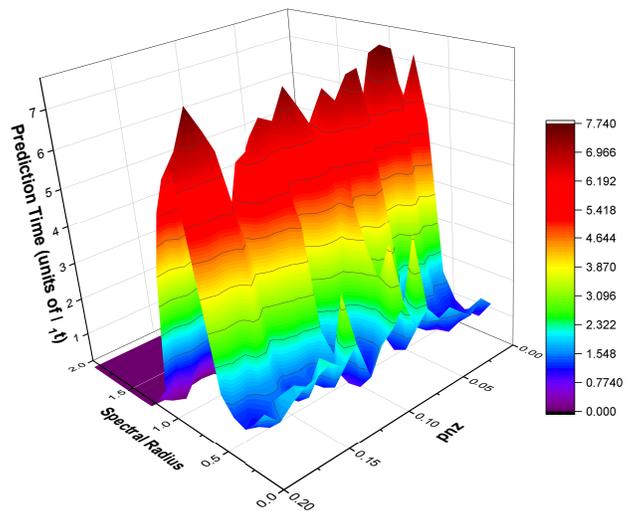}
    \caption{The average forecast time of the reservoir depends strongly on the RC hyperparameters.  Here we show the average forecast time variation (in units of $\lambda_1 t$) as a function of SR and pnz for the Lorenz63 system.  This kind of grid search is very computationally intensive due to the high sensitivity of the RC to changes in the parameters and the necessity of testing at multiple points to ensure stability of the prediction.}
    \label{fig: pred_L63}
\end{figure}

When $\u(t)$ and $\r(t)$ are synchronized, the combined system in $\mathbb{R}^{N+D}$ will lie on a, generally complicated, {\it synchronization manifold}~\cite{Pecora97}. During the training phase when the reservoir evolves according to Eq.\eqref{eq: synchronization_pred}, $\u(t)$ drives the reservoir system towards the synchronization manifold.

GS gives us some advantage in the analysis of RC networks.  GS assures us that the dynamical properties of the stimulus $\u(t)$ and the reservoir $\r(t)$ are now essentially the same. They share global Lyapunov exponents~\cite{Oseledec68}, attractor dimensions, and other quantities classifying nonlinear systems ~\cite{Abarbanel95}.

The principal power of GS in RC is that we may replace the initial non-autonomous reservoir dynamical system with an autonomous system operating on the synchronization manifold. (See Eq.\eqref{eq: synchronization_pred})
 
The function $\varphi(\r)$ is approximated in some manner,  through training, and then this is substituted for $\u$ in the reservoir dynamics. In previous work on this~\cite{Lu18,Hunt19,ottdresden19} the authors approximated $\varphi(\r)$ via a polynomial expansion in the components $\r_{\alpha}$. and used a regression method to find the coefficients of the powers of $\r_{\alpha}$. We follow their example in this paper but note that there is a more general formulation of the problem \cite{Platt_PRE}.

\subsection{The Auxiliary Method for GS}
There are a variety of approaches for determining whether $\r(t)$ and $\u(t)$ exhibit GS. Perhaps the easiest approach is to establish {\bf two identical} reservoirs~\cite{Abarbanel96} driven by the same $\u(t)$, 
\bes
\frac{d\r_A(t)}{dt} = \F_r(\r_A(t), \u(t)) \; \& \; \frac{d\r_B(t)}{dt} = \F_r(\r_B(t), \u(t)).
\ees

Then we compare some function, $\chi(\r)$ of $\r_A(t)$ against the same function of $\r_B(t)$. This should yield a straight line in the $\{\chi(\r_A),\chi(\r_B)\}$ plane. The two states $\r_A(t)$ and $\r_B(t)$ should be identical after a short transient period, even though the initial conditions of the reservoirs are typically different.  This test does not tell us what the function $\varphi(\r)$ is or what any of its properties may be; it only establishes the existence of $\varphi(\r)$.
\begin{figure}[!htpb]
\centering
\includegraphics[width=0.5\textwidth]{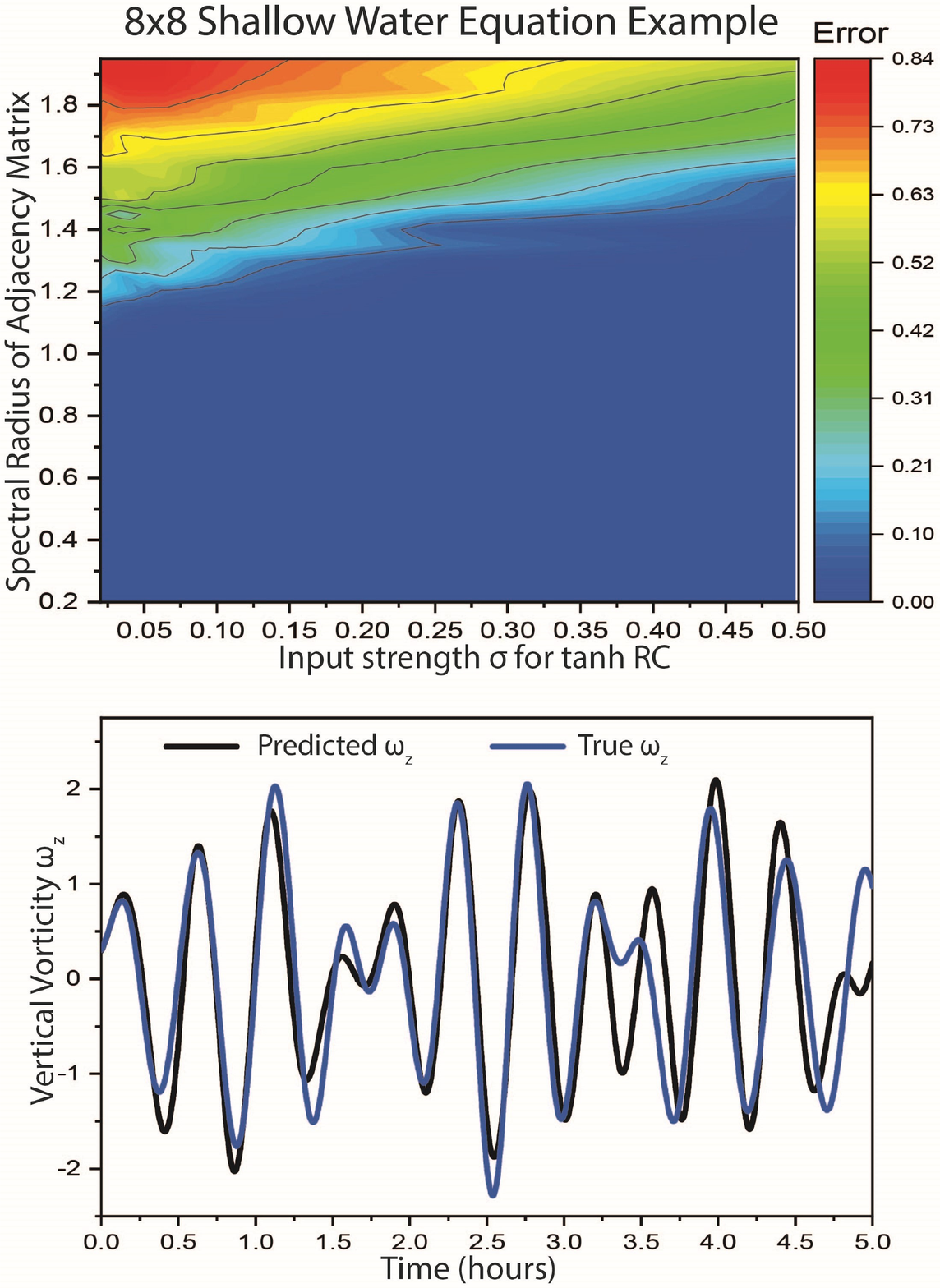}
\caption{\\{\bf Top} Contour plot of \textit{GS} and \textit{no GS} regions. Blue/Purple indicates a region of parameters in a localized $\tanh$ reservoir model (N = 5000) which shows GS with a driving signal from the 8 $\times$ 8 Shallow Water Equations (SWE) \cite{Sadourny75} as $\u(t) \in \mathbb{R}^{192}$. The red region shows no GS.\\ {\bf Bottom} Forecast for the normalized vorticity at a particular point on the 192 dimensional $8\times8 \times 3$  grid.  Localized reservoir scheme and details of the SWE are found in \cite{Platt_PRE}.}
\label{fig: swe}
\end{figure}

\subsection{Synchronization Test}

GS provides us with a test of whether a particular reservoir---with choice of architecture, dynamics and hyperparameters---has the capability to learn the dynamics implied by the data.  Following the previous section, {\bf without training}, one can simply evolve $\F_r(\r(t), \u(t))$ with the input $\u(t)$ present for two different initial conditions, and then test if GS occurs.  If GS {\bf does not occur} between the reservoir and the data then the reservoir is almost certainly untrainable and the choice of hyperparameters needs to be changed---see Fig.(\ref{xlor63tores}).

Looking for GS can greatly reduce the number of RCs with different hyperparameters that must be trained and tested in a traditional grid search---Fig.(\ref{fig: pred_L63}) shows such a grid search. The advantage of searching first for GS comes from the fact that the auxiliary test is fast and efficient.  The conditional Lyapunov exponents \cite{Lyapunov, Pecora90} between the drive and response systems being negative mean that the two reservoir states should converge exponentially towards each other.  In practice this property means that one can look at a much smaller segment of time than is required for accurate training.  In addition, the training step does not need to be completed, so searching for GS is computationally much more efficient than training a reservoir and then evaluating it by predicting at multiple points.

Testing for GS only tells us that the function $\varphi(\r(t))$ exists, not whether our approximation to it is sufficient for prediction.  One would expect a linear approximation to $\varphi(\r(t))$ would predict well only for a small subset of the parameters for which GS is shown to occur; indeed this is exactly what we find empirically.  We hypothesise that more complex approximations might expand this subset of good predictions to include most of the region indicated by the GS test.

An example of using GS to find hyperparameters to predict a geophysical system is shown in Fig. (\ref{fig: swe}).  Here we predict the evolution of the shallow water equations \cite{Sadourny75} on an $8 \times 8$ grid using a localized reservoir scheme described in \cite{Platt_PRE}.  Using the auxiliary method to test for GS, we found a set of parameters for this high dimensional model quite efficiently.

\section{Evaluation of Reservoir Properties} \label{sec: eval}
After running the test for GS and performing a hyperparameter search, the question arises of how to guarantee stable forecasting.  One often encounters the two situations in RC:
\begin{itemize}
    \item The forecast starts out close to the data but then quickly diverges and becomes non-physical
    \item The forecast is ``good'' for certain initial starting conditions but not for others.
\end{itemize}

The problem of finding a set of hyperparameters that will give robust predictions is one of the main challenges in reservoir computing and RNNs in general.  The definition of a robust set of hyperparameters is one in which neither the randomness of the adjacency matrix $A_{\alpha\,\beta}$ or the training data set causes the reservoir to fail in prediction.  An example of robustness is shown in \cite{Platt_PRE} for a Lorenz63 input system.  Many sets of parameters work well for particular inputs. It is more challenging to find a set of parameters that predict well for general input.

\begin{figure}[htpb!]
    \centering
    \includegraphics[width = 0.49\textwidth]{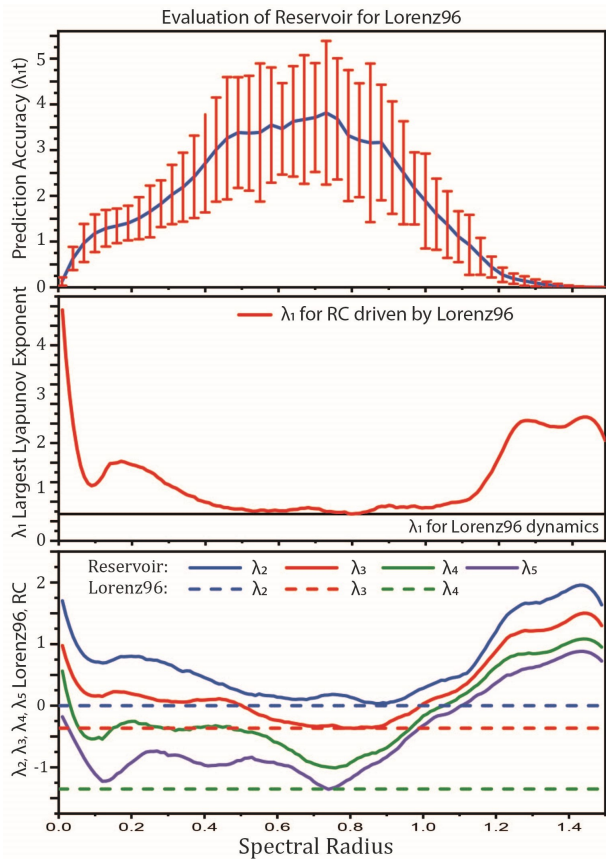}
    \caption{{\bf Top} Average prediction time of a N=2000 $\tanh$ reservoir as a function of SR for $D=5$ Lorenz96 Driver~\cite{Lorenz96, Platt_PRE}. The time units are in $\lambda_1 t$.  The error bars indicate variation in prediction depending on the stability of the input stimulus.
    {\bf Middle} Largest Lyapunov exponent, $\lambda_1$ of the forecast reservoir and the input system (black line) as a function of the spectral radius.
    {\bf Bottom} $\lambda_2,...,\lambda_5$ Smaller Lyapunov exponents for the predicting reservoir. The method for computing the Lyapunov Exponents of an RC is discussed in  ~\cite{abar96,Eckmann85,Verstraeten09,Lu18}.}
    \label{fig: lyap_predlor96}
\end{figure}

The typical approach for evaluating machine learning predictions with the mean squared error over a test set does not capture a key feature of RC.  A well trained RC should be able to give good short term predictions for {\bf all} initial starting points {\bf and} be stable in the medium to long term. This feature is called attractor reconstruction~\cite{Lu18}.  Instead of a test set, we propose an additional criterion for RC evaluation; {\bf a well trained RC reproduces the spectrum of Lyapunov exponents of the input system $F_u$}---see Fig. (\ref{fig: lyap_predlor96}) for an example.

Lyapunov exponents (LEs) characterize the average global error growth rate of a dynamical system \cite{Lyapunov} along directions in phase space.  One can calculate the $N$ LEs of the forecast reservoir Eq.\eqref{eq: synchronization_pred} and compare them to the $D$ exponents of the input system.  If the $D$ largest LEs match and the smaller $N-D$ exponents of the RC are negative, then the two systems will have the same global behavior, increasing the likelihood of robust, stable predictions.

We show this calculation for the Lorenz96 system~\cite{Lorenz96} in Fig.(\ref{fig: lyap_predlor96}). Our results show that when more of the spectrum of LE's are matched by the RC, the better the average predictions. If the Lyapunov spectrum of the RC does not match that of the input then the two situations above are more likely to occur.  In situations where it is difficult to exhaustively test the RC, perhaps because the model is expensive to run or there is limited data, evaluating the Lyapunov exponents of the reservoir will guarantee that the global error growth of the RC is the same as the data.  A similar calculation is performed in \cite{Lu18} but without systematically tying the results to the average prediction time.

The results presented in Fig.(\ref{fig: lyap_predlor96}) match the suggestion that the reservoir operates best at ``the edge of chaos''~\cite{Verstraeten09, Boedecker12, Jiang19}, that is, the maximal prediction time of the reservoir corresponds to a SR just less than 1.  We make the case that the ``edge'' corresponds to the state where the reservoir LEs approximately match all the non-negative exponents of the input system.

\section{Conclusions}
Recurrent neural networks are a powerful tool for time series prediction tasks.  While much intuition and knowledge for practical applications have been built up for specific tasks over the years, understanding the tradeoffs when designing a particular network is of the utmost importance.  In this paper we
\begin{enumerate}
\item Introduced a test based on the property of GS that helps narrow down the hyperparameter search space when designing an RNN for a specific problem
\item Explained the connection between GS and RCs, which led us to be able to gain insight into how the properties (LEs) of the synchronization manifold affect the ability of the RC to predict accurately forward in time, thus enabling us to set new evaluation criteria for RC
\end{enumerate}

We have explored the role of GS, where the input $\u(t)$ driving the reservoir and the reservoir coordinates $\r(t)$ satisfy $u_a(t) = \varphi_a(\r(t))$. We have elaborated on the notion that the only training required to provide accurate estimations/forecasts by the trained reservoir involves the estimation of parameters in representations of $\varphi(\r)$.

%As a final comment, let us recall that for the use of %these methods in a practical use of forecasting with %RC, it seems likely we will need to implement the %reservoir in hardware for computational %efficiency~\cite{canaday18}.

 \section*{Acknowledgments} We had many thoughtful discussions on the material presented here with Brian Hunt. Three of us (SGP, J. Platt, and HDIA) acknowledge support from the Office of Naval Research (ONR) grants N00014-19-1-2522 and N00014-20-1-2580. SGP acknowledges further support from the National Oceanographic and Atmospheric Administration (NOAA) grants NA18NWS4680048, NA19NES4320002, and NA20OAR4600277. We also thank Lawson Fuller for many fruitful discussions and for paper editing.

%\bibliography{PRL}
\end{document}